\documentclass[
]{ceurart_preprint}

\usepackage{listings}
\usepackage{url}
\usepackage{todonotes}

\usepackage{caption} 
\usepackage{subcaption} 
\clearpairofpagestyles
\begin{document}

\copyrightyear{2024}
\copyrightclause{Copyright for this paper by its authors.
  Use permitted under Creative Commons License Attribution 4.0
  International (CC BY 4.0).}

\conference{D2R2’24: Third International Workshop on Linked Data-driven Resilience Research 2024}

\title{Leveraging small language models for Text2SPARQL tasks to improve the resilience of AI assistance}

\author[1]{Felix Brei}[%
orcid=0009-0008-5245-6655,
email=brei@infai.org
]
\cormark[1]

\author[2,3]{Johannes Frey}[%
orcid=0000-0003-3127-0815,
email=frey@informatik.uni-leipzig.de,
url=,
]

\author[1,3]{Lars-Peter Meyer}[%
orcid=0000-0001-5260-5181,
email=lpmeyer@infai.org
]

\address[1]{ETi Competence Center @ Institute for Applied Informatics, Germany, \url{https://cc-eti.org}}

\address[2]{KMI Competence Center @ Institute for Applied Informatics, Germany, \url{https://kmi-leipzig.de}}

\address[3]{Institute of Computer Science, Leipzig University, Germany, \url{https://cs.uni-leipzig.de} }

\cortext[1]{Corresponding author.}

\begin{abstract}
In this article we show that small language models with less than one billion parameters can be fine-tuned to be able to translate natural language to SPARQL queries with a precision that comes close to those of large language models like ChatGPT, OpenLlama and others. We will identify the prerequisites for the training of these models and what to look out for. The goal of this research is to enable people to fulfill these translation tasks on affordable commodity hardware as well as reducing the environmental impact that comes with the use of LLMs.
\end{abstract}

\begin{abstract}
In this work we will show that language models with less than one billion parameters can be used  to translate natural language to SPARQL queries after fine-tuning. Using three different datasets ranging from academic to real world, we identify prerequisites that the training data must fulfill in order for the training to be successful. The goal is to empower users of semantic web technology to use AI assistance with affordable commodity hardware, making them more resilient against external factors.
\end{abstract}


\maketitle

\section{Introduction}

    The usage of Large Language Models (LLMs) has increased exponentially since the advent of ChatGPT. According to Similarweb, the website of OpenAI alone was visited more than 1.6 billion times by February 2024\footnote{\url{https://www.similarweb.com/website/openai.com/\#overview}}. In addition to that, Microsoft has launched several AI assistants called 'Copilots' which are based on LLMs \footnote{\url{https://blogs.microsoft.com/blog/2023/09/21/announcing-microsoft-copilot-your-everyday-ai-companion/}}$^,$\footnote{\url{https://github.blog/2023-11-08-universe-2023-copilot-transforms-github-into-the-ai-powered-developer-platform/}}, as well as Google releasing their AI called Bard which is now known as Gemini\footnote{\url{https://blog.google/technology/ai/bard-google-ai-search-updates/}}$^,$\footnote{\url{https://blog.google/technology/ai/google-gemini-ai/}}. This suggests that the big tech companies believe in the potential of LLMs to become part of our daily lives, just like smartphones or computers in general. But do they hold up to the expectations?

    Several test suites were derived to assess the generative capabilities of LLMs, for example TruthfulQA\cite{lin2022truthfulqa}, HellaSwag \cite{zellers2019hellaswag} or the Abstraction and Reasoning Corpus (ARC) \cite{chollet2019measure}. These test suites, among others, are run regularly on the latest entries to the LLM circus and the results for open LLMs are presented publicly on the Huggingface OpenLLM leaderboard \cite{open-llm-leaderboard}. We can see that the performance increases drastically over time, with Bloom \cite{workshop2023bloom} scoring an average of 46.06 in August 2023 and Smaug-72B\footnote{\url{https://huggingface.co/abacusai/Smaug-72B-v0.1}} holding the record in February 2024 with a score of 80.48, only half a year later.

    These test suites however cover mostly natural language domains, like the task to continue a sentence, answer questions or extract information from a paragraph of text. Based on the experience from early experiments \cite{Meyer2023LLMassistedKnowledge}, a test suite \cite{meyer2023developing} was developed that evaluates capabilities of LLMs to interface knowledge graphs and assist in knowledge engineering tasks. 
    While the smaller open-source GPT4All models severely struggled, the state-of-the-art commercial LLMs GPT4 and Claude showed promising results \cite{DBLP:conf/dl4kg/FreyMABB23} and a trend of performance improvements over the course of 2023 \cite{frey-llm-evo} in dealing with KGs in Turtle format. 

    Alas, these results come with several caveats:

    \begin{enumerate}
        \item The commercial LLMs that were tested are all hosted externally. This can be problematic regarding data protection, because a user has to send a information to a third party.
        \item Because of their sheer size (GPT4 has one trillion parameters\footnote{\url{https://www.semafor.com/article/03/24/2023/the-secret-history-of-elon-musk-sam-altman-and-openai}}), running these models locally is prohibitively costly and therefore not an option for a lot of research institutes and other parties. On top of that, training a model of such size is also extremely expensive\footnote{\url{https://www.wired.com/story/openai-ceo-sam-altman-the-age-of-giant-ai-models-is-already-over/}}.
        \item
        Even these commercial models were at the time of writing still significantly challenged by SPARQL query generation or RML mapping generation    \cite{DBLP:conf/dl4kg/FreyMABB23,hofer2022towards} indicating a need for specific training or fine-tuning of all models w.r.t. handling those tasks in a reliable and efficient way.
        \item Since all these larger are hosted on third party platforms, users are at the mercy of the vendors to keep the services running and affordable. However, vendors suddenly changing their licensing and cost model has already happened in the past\footnote{\url{https://www.theverge.com/2023/9/12/23870547/unit-price-change-game-development}}, as well as deep sea cables being damaged\footnote{\url{https://edition.cnn.com/2024/03/04/business/red-sea-cables-cut-internet/index.html}}, separating certain areas of the world from the internet and leaving local companies only with the computational resources they have on site.
    \end{enumerate}

    So we ask ourselves the following question: Given a single task that we want so solve using LLMs,
    is it possible to achieve a similar performance of these large models with a much smaller one? This would enable small businesses to use AI assistance with affordable hardware they can host on site, increasing their resilience against outages, vendors changing their pricing models, disruption due to trade embargoes and other external factors.

    As a first step into this direction, we study the task of translating a natural language question into a SPARQL query because we think that this task enables people who are not familiar with SPARQL to extract knowledge and insights from a knowledge graph which would otherwise not be possible for them. 
    The paper is organized as follows: First, we look at related research in this field and explain where we fit into the big picture. Then we explain the setup of our experiments, namely which model families were chosen and why and which datasets we trained them on. After that, we present and explain the results of our work and finally, we draw conclusions and give an outlook on the directions that our research will head next.

\section{Related Work}
    
    Current approaches focus on fine-tuning large language models. For example the authors of  \cite{rangel2024sparql} propose a methodology for fine-tuning OpenLLaMA to generate SPARQL queries over life science knowledge graphs using data augmentation techniques, such as providing meaningful variable names and inline comments, improving the performance of the model in generating accurate SPARQL queries. The authors of \cite{xu-etal-2023-fine} use Llama as their basis for fine-tuning to generate SPARQL queries over Wikidata.

    These two papers have shown that translating natural language to SPARQL queries is possible, but they use models with at least three (OpenLLaMA) resp. seven (LLaMA) billion parameters.
    The hardware required to train these models can be expensive, which is why we want to explore models that are even smaller.

    Smaller, fine-tuned models for one specific task are also able to beat the performance of LLMs, e.g. SQLCoder-7B \footnote{\url{https://huggingface.co/defog/sqlcoder-7b-2}} performs better on SQL than state of the art GPT4. Our research is comparable to that, but with much less parameters and SPARQL instead of SQL.

    \cite{Banerjee_2022} manages to fine-tune T5 on SPARQL queries for Wikidata, but to achieve these results, the data had to be preprocessed in a way that is specific to T5. Furthermore, while this paper explores other ways to tackle this task in general, it only looks at T5 instead of other model families as we do.
    
    \cite{diallo2024comprehensive} gives a comprehensive overview and performs a comparison of pre-trained LMs (PLMs), non-pre-trained LMs (NPLMs), and LLMs, testing various fine-tuning methods using LLMs.
     \cite{Li2023FlexKBQAAF}
     fine-tunes a lightweight model for SPARQL generation using synthetic training data generated by the FlexKBQA framework on a target knowledge graph (sampling structured query templates that are converted into SPARQL query instances and translated into natural language questions using LLMs).
     The light-weight model can perform further self-guided training on real queries to address a distribution shift between synthetic and real queries.    
    \cite{Bustamante2024SPARQLGW} uses a GPT model to investigate what parts of the Text2SPARQL task are the hardest for the model to solve so appropriate countermeasures can be taken.
    
    \cite{9815253} proposes a whole new architecture specific for SPARQL generation based on GPT. This direction we assume promising for the future, but here we are focusing on more foundational research first to understand which model families work best on a given dataset and why.

\section{Experimental Setup}

\subsection{Model families}

As was mentioned in the introduction, the focus of our work is to fine-tune language models that can be considered small by modern standards. We chose one billion parameters as an arbitrary limit on the number of parameters, but as a general guideline we consulted the Steam Hard- and Software Survey\footnote{\url{https://store.steampowered.com/hwsurvey/}} and found that $57.22\%$ of their users use a GPU with 8GB of VRAM or more (January 2024). A model with less than one billion parameters should fit into this amount of VRAM comfortably, showing again that these LLMs can be trained and run locally.

Another consideration is the public availability of the models. 
We believe that research should be available to anyone who is interested and this should be reflected in the choice of models. 
Therefore, we only select models that are openly available on Huggingface.

Following these criteria, we observe quickly that there are only three large model families that fit the bill, which we introduce here briefly. A full list of models evaluated is given in table \ref{tab:model_selection}

\subsubsection{T5 and Flan-T5}

In June 2020 Google released an LLM called Text-To-Text Transfer Transformer, or T5 in short \cite{2020t5}. The base version consists of roughly 220 million parameters, with smaller and larger versions available. With T5, Google wanted to provide a single LLM that can solve any NLP task like text classification, sentiment analysis and so on. A user must provide a prefix like '\textit{Translate the following sentence to french:}' and the LLM then infers how to process the rest of the prompt. In 2022, researchers at Google released new versions of T5 called FLAN-T5 \cite{2022flant5} (FLAN stands for \textit{fine-tuning language models} \cite{wei2022finetuned}) which, according to the authors, should outperform T5 on any given task.

\subsubsection{BART}

BART was developed by Facebook and released in October 2019 \cite{bart}. It consists of 139 million parameters and is a combination of a BERT-like encoder \cite{devlin2019bert} with a GPT-like autoregressive decoder \cite{radford2019language}. In August 2020, a multilingual version called mBART was released \cite{tang2020multilingual}. The authors put special emphasis on the fact that BART is just a pretrained model and needs to be fine-tuned for a given specific task. We also included mREBEL models as a specialized version of BART for multilingual relation extraction \cite{huguet-cabot-et-al-2023-redfm-dataset} since it was finetuned with knowledge graphs in mind.

\subsubsection{M2M100 and NLLB-200}

The M2M100 model was introduced in 2020 \cite{fan2020englishcentric} as a many-to-many translation tool for 100 languages. The original version consists of 1.3 billion parameters which exceeds the upper bound we imposed. But there is a distilled version available directly from the Facebook research team at Huggingface called M2M100-418M\footnote{\url{https://huggingface.co/facebook/m2m100_418M}} which we use in our experiments.

Its successor, the NLLB-200 model, was introduced in 2022 \cite{nllbteam2022language} and stands for \textit{'no language left behind'}. Again we use the distilled version NLLB-200-Distilled-600M\footnote{\url{https://huggingface.co/facebook/nllb-200-distilled-600M}} instead of the 3.3 billion full version of the model. As the authors state, the model is \textit{'primarily intended for research in machine translation'} which fits our bill perfectly.

This leaves us with a selection of models to be assessed in our experiment that can be seen in table \ref{tab:model_selection}.

\begin{table}[tb]
    \centering
    \caption{Model names and their number of parameters, as used in our experiments.}
    \label{tab:model_selection}
    \begin{minipage}{0.45\linewidth}
        \centering
        \begin{tabular}{c|c}
            \textbf{Name} & \textbf{No. parameters} \\ \hline
        T5-Small  & 60.5M \\
        T5-Base  & 223M \\
        T5-Large  & 738M \\
        FLAN-T5-Small  & 77M \\
        FLAN-T5-Base  & 248M \\
        FLAN-T5-Large  & 783M \\
        \end{tabular}
    \end{minipage}%
    \hfill
    \begin{minipage}{0.45\linewidth}
        \centering
        \begin{tabular}{c|c}
            \textbf{Name} & \textbf{No. parameters} \\ \hline
        BART-Base  & 139M \\
        BART-Large  & 406M \\
        mBART-LARGE-50  & 611M \\ 
        mREBEL-Base & 484M \\ 
        mREBEL-Large & 611M \\ \hline
        M2M100-418M & 418M \\
        NLLB200-Distilled-600M & 600M
        \end{tabular}
    \end{minipage}
\end{table}

\subsection{Datasets used for Fine Tuning and Evaluation}
In order to study how well the models can be fine-tuned towards a target KG, we use three evaluation datasets from different domains and with varying complexity.
These datasets are comprised of a number of natural language questions, which are mapped to a SPARQL query w.r.t. the target KG.
For the first two datasets (organizational graph and CoyPu graph) we generate questions and queries by sending the graph via the OpenAI API to GPT4 and prompting it to generate tuples of natural language question, matching SPARQL query, and the expected result of the query.
These tuples are filtered by checking if the results that the SPARQL query returns match with the expected results. 
Both datasets are then augmented by sending each remaining question again to GPT and asking it to paraphrase the question, giving us a total of two natural language questions per SPARQL query.

\subsubsection{Organizational Graph}

Introduced in \cite{Meyer2023LLMassistedKnowledge}, this small knowledge graph uses established vocabularies to describe an organization with departments and employees. There is a clear schema that maps person and department names to their corresponding RDF resource, for example \textit{"Anne Miller"} maps to \verb|:anne| while \textit{"Bob Tanner"} maps to \verb|:bob|. In this dataset and the next we also let the language model omit the prefix definitions for the queries and assume they are already present in the preamble of the executed SPARQL query. Using GPT4 we generated a dataset consisting of 69 datapoints, which were split into 53 tuples for training and 16 for testing.

\subsubsection{A subset of the CoyPu graph}

The CoyPu project\footnote{\url{https://coypu.org/}} aims to improve supply chain resiliency for corporations by combining different data sources about public infrastructure, trades and trade agreements, events like disasters and conflicts and many more into a large knowledge graph. Querying this knowledge graph has the potential to help businesses identify risks like single points of failures and mitigate them. This usefulness combined with the fact that the other two datasets have more of an academic background made us decide that we use a subset of the CoyPu knowledge graph as another dataset for training. Creating a viable subset lead to its own difficulties and hurdles however, which we consider as future work. This dataset contains 131 tuples in total, which were split into 105 for training and 26 for testing.

\subsubsection{QALD10}

The \textit{Question Answering over Linked Data} (QALD) dataset is a standard benchmark\footnote{\url{https://www.nliwod.org/challenge}} with QALD10 being the latest incarnation \cite{qald10}. It consists of SPARQL queries along with matching questions in different natural languages, w.r.t. Wikidata. 
In this work, we focus on English and filter the dataset accordingly. This dataset is especially difficult for a language model to handle because there is no clear indication how to link entities from a given question like \textit{"Barack Obama"} to their Wikidata entity ID (\verb|:Q76|), giving rise to a whole field of research called Entity Linking \cite{entity}.

\subsection{Fine-tuning}
For every evaluation dataset individually, we perform fine-tuning of the selected models using PyTorch (100 epochs).
Since a single run of fine-tuning does not hold much statistical significance and involves random parameters, we performe isolated runs of the training for a total of ten times.
For each run we shuffle the training data with a predetermined random seed to make the results reproducible. 
Specifically, each run is given an ID from $R01$ to $R10$ and the seeds are generated by calculating the SHA512 sum of the ID and taking the first eight digits, so $R01$ results in the seed $99975818$, $R02$ in $56899599$ and so on.

\section{Results}

In the following subsections we only include those language models in the plots that generated at least one correct query. The T5 family consistently generated not a single correct query on the organizational graph which is why it is absent in the result tables and figures. In fact, all T5 models did not produce a single correct result across all runs. 

To generate the datapoints for each plot, we interrupted the training every five epochs and made the language models translate the questions from the evaluation split into SPARQL queries. 
We then executed the queries and compared the result sets to determine whether the answers is correct.

\subsection{Organizational Graph}

\begin{figure}
    \centering
    \includegraphics[width=\columnwidth]{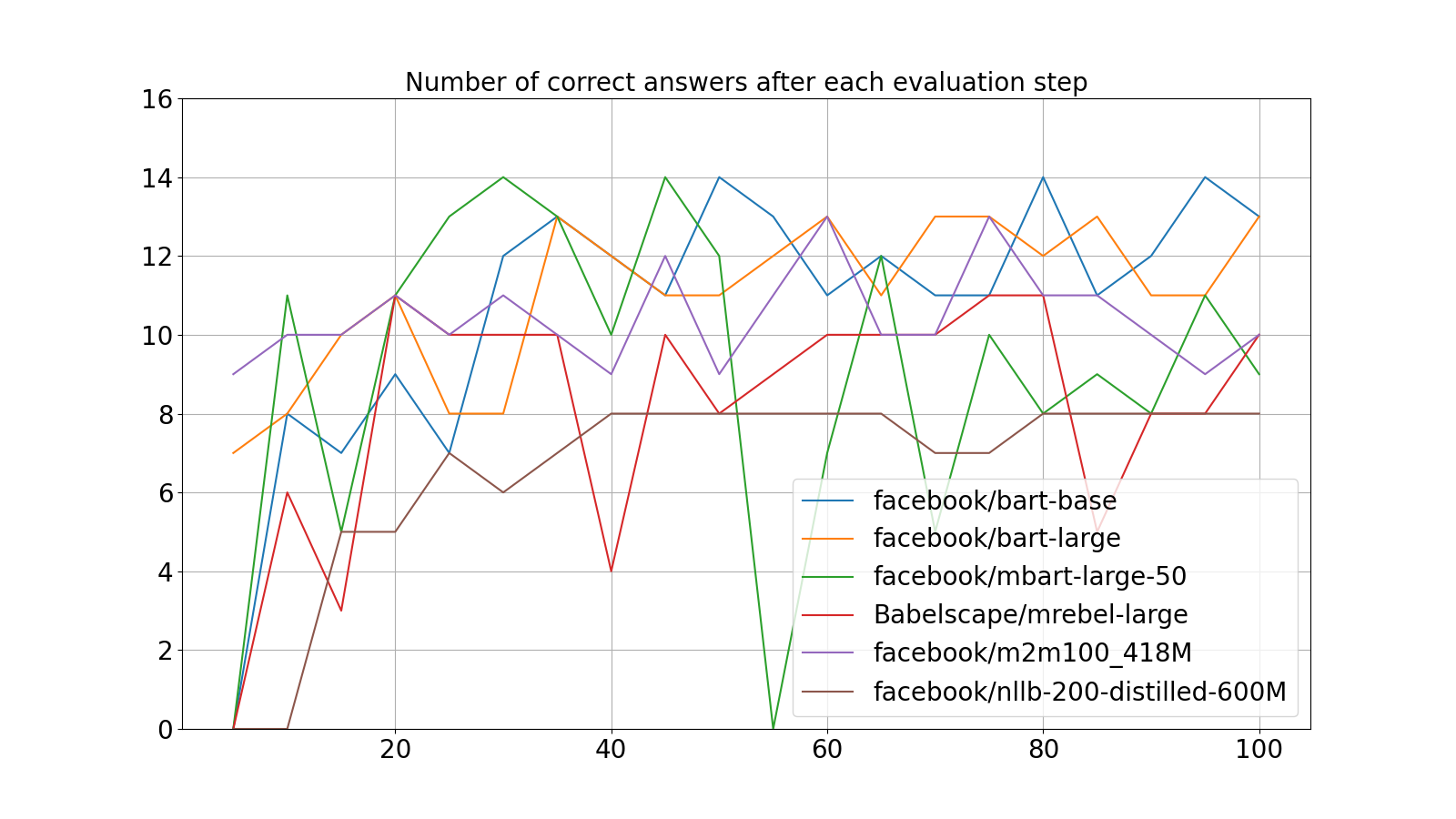}
    \caption{Number of correct SPARQL translations across epochs for the organizational graph for one fine-tuning run per model. 16 questions were presented.}
    \label{fig:orga-all}
\end{figure}

Figure \ref{fig:orga-all} shows that all models from the BART and M2M100 families manage to learn the structure of the knowledge graph at least to a certain degree. When taking the best results for each model, aside from NLLB-200, all models turn at least eleven of the sixteen questions into correct SPARQL queries. The performance however fluctuates extremely during the course of the training which is indicative of overfitting.

\begin{figure}
    \centering
    \begin{subfigure}{0.49\columnwidth}
        \includegraphics[width=\columnwidth]{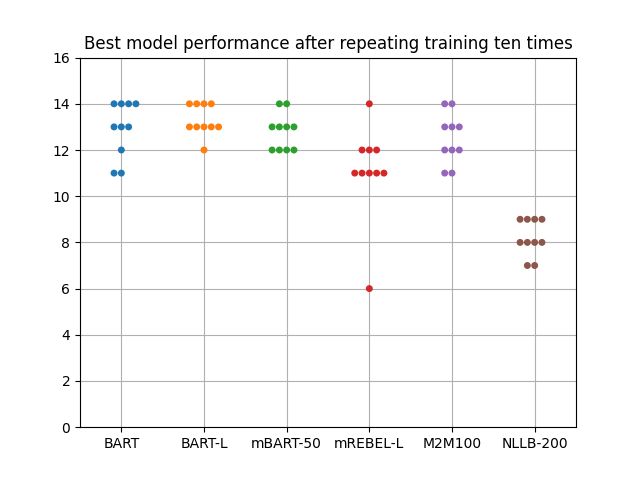}
        \caption{Organizational Graph}
        \label{fig:boxes_orga}
    \end{subfigure}
    \begin{subfigure}{0.49\columnwidth}
        \includegraphics[width=\columnwidth]{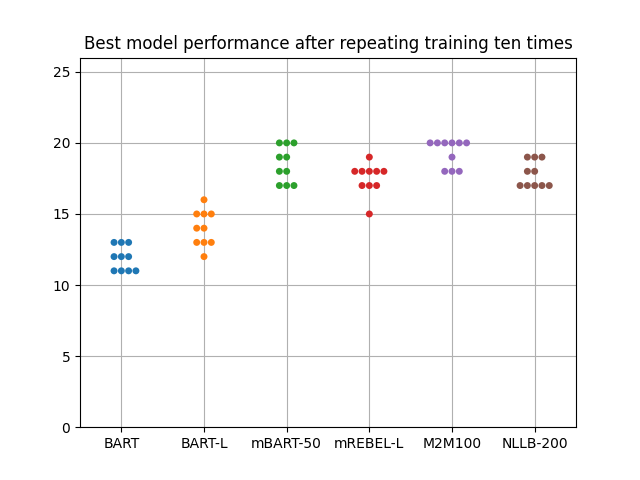}
        \caption{CoyPu dataset}
        \label{fig:boxes_coypu}
    \end{subfigure}
    \caption{Repeating the experiment ten times for both datasets then selecting the best model from each run}
    \label{fig:boxes}
\end{figure}

Repeating the experiment we can see that performance varies a bit depending on the order the training data is ingested into the network. The statistics are shown in table \ref{tab:stats_orga} and the raw data is plotted on the left side of figure \ref{fig:boxes}. We can see that for this dataset, BART-L performs best (as well as the other sizes of BART), with M2M100 being close behind. Another thing we see from the left plot in figure \ref{fig:boxes} is that except for one outlier coming from mREBEL-L, the success of fine-tuning is reliable and reproducible.

Looking at common errors made during the translation, we found that the best models rarely generated SPARQL that could not be parsed, but they rather mixed up terms and injected parts of the training data into the queries. An example is shown in table \ref{tab:orga_error_example}.

\begin{table}[]
    \caption{An example of the errors that were made in the context of the organizational graph: no binding variable, wrong entity ID :charles, wrong property foaf:firstName, wrong literal BobTanner }
    \label{tab:orga_error_example}
    \centering
    \begin{tabular}{r|l}
        \textbf{Question} & What is the surname of Bob Tanner? \\
        \textbf{Gold answer} & \verb|SELECT ?surname WHERE { :bob foaf:surname ?surname . }| \\
        \textbf{Generated query} & \verb|SELECT ?surname WHERE { :charles foaf:firstName 'BobTanner' }|
    \end{tabular}

\end{table}

\begin{table}[]
    \caption{Average number of correctly translated questions in the context of the organizational graph. Standard deviation is also shown as a percentage of the average to measure the reliability of the fine-tuning.}
    \label{tab:stats_orga}
    \centering
    \begin{tabular}{|l|c|c|c|}
    \hline
    \textbf{Model name} & \textbf{Average} & \textbf{Standard deviation} & \textbf{Std. dev percent} \\
    \hline
BART       & 12.90 & 1.14 & 8.80 \\
\textbf{BART-L}     & \textbf{13.30} & \textbf{0.64} & \textbf{4.81} \\
mBART-50   & 12.80 & 0.75 & 5.85 \\
mREBEL-L   & 11.10 & 1.92 & 17.31 \\
M2M100     & 12.50 & 1.02 & 8.20 \\
NLLB-200   & 8.20 & 0.75 & 9.13 \\
\hline
    \end{tabular}

\end{table}

\subsection{CoyPu}

\begin{figure}
    \centering
    \includegraphics[width=\columnwidth]{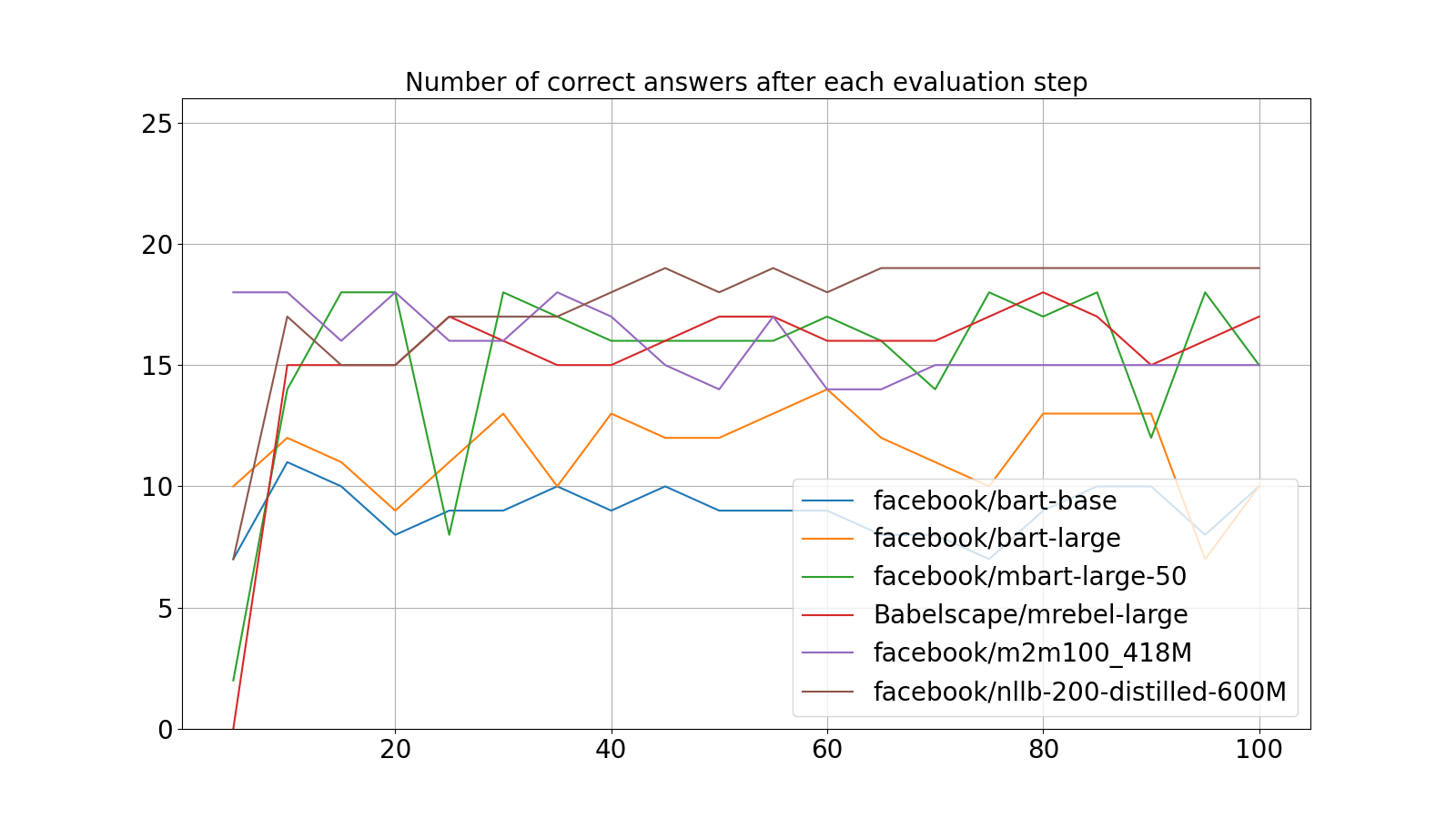}
    \caption{Number of correct SPARQL translations across epochs for the CoyPu dataset for one fine-tuning run per model. 26 questions were presented.}
    \label{fig:coypu-all}
\end{figure}

In figure \ref{fig:coypu-all} we can see that the performance during the first run of the experiment varies less drastically than for the Organizational Graph. The standard deviations seen in table \ref{tab:stats_coypu} are similar though so we think this is just a coincidence. Again, the (FLAN-)T5 models never generate even a single correct query, so they are excluded from consecutive runs.

We can also see that for this dataset, M2M100 outperformed the other models and BART-L is in fact one of the worst, which is a complete shift from the results before. This again shows that one should always evaluate more than a single model since the performance seems to be tied to the structure of the underlying data for fine-tuning.

Again BART-L did not generate a lot of parsing errors, but instead mixes up terms from the supply chain domain, see for example table \ref{tab:coypu_error_example_1}.

\begin{table}[]
    \caption{An example of the errors that were made in the context of the CoyPu graph: latitude and longitude got mixed. \\
    The ellipsis on the IRI was inserted by ourselves to keep the line shorter. In fact, the language model generated the correct IRI to use in the query.}
    \centering
    \begin{tabular}{r|l}
        \textbf{Question} & What is the latitude of the port with the ID 'AUDKB'? \\
        \textbf{Gold answer} & \verb|SELECT ?latitude WHERE { <https://data.coypu.org/...| \\
        & \verb|ns2:hasLatitude ?latitude }| \\
        \textbf{Generated query} & \verb|SELECT ?latitude WHERE { <https://data.coypu.org/...| \\
        & \verb|ns2:hasLatitude ?longitude }|
    \end{tabular}
    \label{tab:coypu_error_example_1}
\end{table}

On the other hand, the main errors that M2M100 made were generating SPARQL queries that could not be parsed (i.e. grammatically incorrect).

\begin{table}[]
    \caption{Average number of correctly translated questions in the context of the CoyPu graph. Standard deviation is also shown as a percentage of the average to measure the reliability of the fine-tuning.}
    \label{tab:stats_coypu}
    \centering
    \begin{tabular}{|l|c|c|c|}
    \hline
    \textbf{Model name} & \textbf{Average} & \textbf{Standard deviation} & \textbf{Std. dev percent} \\
    \hline
BART       & 11.90 & 0.83 & 6.98 \\
BART-L     & 14.00 & 1.18 & 8.45 \\
mBART-50   & 18.50 & 1.20 & 6.51 \\
mREBEL-L   & 17.50 & 1.02 & 5.86 \\
\textbf{M2M100}     & \textbf{19.30} & \textbf{0.90} & \textbf{4.66} \\
NLLB-200   & 17.80 & 0.87 & 4.90 \\
\hline
    \end{tabular}
\end{table}

\subsection{QALD10}

The language models have a really hard time with the QALD10 dataset. While the structure of the generated query comes close to the correct ones, the models cannot handle the translation from entity names like \textit{Kobe Bryant} to their corresponding Wikidata IDs like \verb|Q25369|. We expected this to happen since the whole field of entity linking is ongoing research and far from trivial.

Another problem here is that the QALD10 dataset requires the inclusion of all necessary prefix definitions as part of the query, which was not a requirement both in the organizational as well as the CoyPu graph datasets.

To provide some numbers for clarification: The best performing model was M2M100-418M. The validation dataset contains 394 questions, and only 104 were turned into SPARQL queries that could be parsed. Out of these 104 parsable questions, 51 returned an empty result set. The remaining queries except for three used \verb|COUNT| and returned $0$ because the result set of the underlying query was empty. The three final ones did return wrong bindings.

BART-L only managed to translate one single question into a valid SPARQL query, but the result set was not correct. Interestingly, mBART-L generated 101 parsable SPARQL queries, which makes it a close second to M2M100-418M. The error distribution is about the same as for M2M100-418M though, so no question was correctly answered.

\section{Conclusion and Future Work}

In this paper we have shown that fine-tuning language models for the translation of natural language to SPARQL queries is indeed possible, although there are some limitations like the requirement of a clear and concise mapping from entities in a question to entities in the knowledge graph, like \textit{Anne Miller} to \verb|:anne| instead of \verb|:person1234|. If this requirement is met, both the BART family as well as the M2M100 family are able to fulfill this task.

There is a large amount of avenues that can be explored from here on. First, we should find a better way to define the limit on the number of parameters that a model is allowed to have. Here we have focused on maximum one billion, as this is a limit for most consumer GPUs, but probably there is a connection between model size and sparql generation capabilities.

Secondly, we want to explore how these results can be used to deploy a fine-tuned language model next to a RAG agent to improve its question answering capabilities. So far, LLMs used by RAG agents often lack the ability to correctly apply aggregate functions, which could be remedied by offering the RAG agent a SPARQL query as another source of information.

Since all these models are open source, we can also modify them by manipulating existing layers as well as removing some or inserting new ones. This might be a way to reduce inference time and improve the performance even further. One could also derive completely new models from scratch, since most pre-training datasets are openly available and pre-training is fast due to the small size of the models.

And on top of that, we still have the problem that both the organizational graph dataset as well as the CoyPu dataset were generated using GPT which defeats the purpose of being independent from third parties. We will also investigate in the future how the training data can be generated with open source LLMs like Falcon, Bloom and others so even this step of the pipeline can be executed locally. Here it does not matter if we have enough GPU memory available, since the creation of the training and testing datasets is only done once, so it is not an issue if this step takes a bit longer.

The goal of this paper was to do a small survey of the out-of-the-box capabilities of readily available language models. What we have seen so far looks promising and there is a lot of intriguing research to be done in the near future.


\newpage
\begin{acknowledgments}
This work was partially supported by grants from the German Federal Ministry for Economic Affairs and Climate Action (BMWK) to the CoyPu project (01MK21007A) and KISS project (01MK22001A) as well as from the German Federal Ministry of Education and Research (BMBF) to the project StahlDigital (13XP5116B).
\end{acknowledgments}

\bibliography{refs}

\appendix

\section{Online Resources}

Source code for the training, organizational graph dataset and CoyPu dataset can be found at \href{https://github.com/AKSW/LMs4Text2SPARQL}{https://github.com/AKSW/LMs4Text2SPARQL} and at \href{https://zenodo.org/doi/10.5281/zenodo.10996425}{DOI:10.5281/zenodo.10996425}.

\end{document}